%% file: neurips_2026.tex
\documentclass{article}

\usepackage[preprint]{neurips_2026}

\usepackage[utf8]{inputenc}
\usepackage[T1]{fontenc}
\usepackage{wrapfig}
\usepackage{hyperref}
\usepackage{url}
\usepackage{booktabs}
\usepackage{amsfonts}
\usepackage{amsmath}
\usepackage{amssymb}
\usepackage{mathtools}
\usepackage{adjustbox}
\usepackage{amsthm}
\usepackage{nicefrac}
\usepackage{microtype}
\usepackage{xcolor}
\usepackage{graphicx}
\usepackage{subcaption}
\usepackage{multirow}
\usepackage{enumitem}

\usepackage[capitalize,noabbrev]{cleveref}

\usepackage[textsize=tiny]{todonotes}

\theoremstyle{plain}

\theoremstyle{definition}

\theoremstyle{remark}

\title{
EvtGraph: Event-Adaptive Compression for Sparse Temporal Graph Learning in Multimodal Time Series
}

\author{%
  Ziqian Wang \\
  Tingxiong Xiao \\
  Yuxiao Cheng \\
  Jinli Suo \\
  Department of Automation \\
  Tsinghua University \\
  Beijing, China \\
  \texttt{\{wangziqi24, xtx22, cyx22, jlsuo\}@mail.tsinghua.edu.cn}
}

\begin{document}

\maketitle

\begin{abstract}
Multimodal temporal data are inherently irregular and uneven in information density, 
yet most models rely on uniform discretization, leading to inefficient representations. 

We propose \textbf{EvtGraph}, a unified framework that aligns computation with temporal salience under explicit budget constraints. 
EvtGraph reparameterizes sequences into event-level tokens via event-adaptive compression (EAMC), 
selects a compact subset with a node budget (NBC), 
and performs temporally constrained sparse graph reasoning (T2SG). 
This transforms dense sequences into structured computation over salient events, 
reducing complexity while preserving critical transitions.

We show that this design provides a practical mechanism for allocating representational capacity under a fixed budget, 
yielding a consistent performance--efficiency trade-off, where a small budget is often sufficient in practice.
Experiments on multimodal clinical (MIMIC-IV + CXR) and cross-domain benchmarks 
demonstrate that EvtGraph outperforms both Transformer-based and recurrent baselines 
while significantly improving efficiency.

These results suggest that budget-constrained event-centric representation 
provides a general paradigm for learning from high-redundancy temporal data.
\end{abstract}

\section{Introduction}
\label{introduction}

Multimodal temporal data are ubiquitous across real-world applications, ranging from clinical monitoring with physiological signals and medical records to video understanding that integrates visual, auditory, and textual streams. A fundamental challenge in modeling such data lies not only in heterogeneous sampling rates or long-range dependencies, but in a deeper structural mismatch: while modern architectures allocate computation uniformly over time, real-world temporal signals are inherently non-uniform in information density. Informative patterns are sparse, irregular, and concentrated in short intervals, whereas large portions of the sequence are redundant.

This mismatch induces systematic inefficiencies. Uniform discretization schemes, such as fixed windows or patch-based representations, over-allocate representational capacity to low-information regions while fragmenting semantically coherent temporal structures. When lifted into graph representations for multimodal reasoning, this inefficiency further manifests as the \emph{node explosion} problem, where the number of nodes grows linearly with sequence length across modalities and dense connectivity leads to quadratic complexity, making standard graph neural networks impractical for long sequences.

Existing approaches partially alleviate this issue but remain limited by their underlying assumptions. Multi-scale or pyramid-based models \citep{li2021pyraformer, wu2023timesnet} retain predefined temporal grids and thus fail to align resolution with information density. Sparse attention and correlation-based graph methods \citep{zaheer2020bigbird, zhang2020mmgnn} reduce computational cost but operate on fixed tokenizations, leaving the representation itself unchanged. Token pruning and pooling strategies \citep{bolya2023tome} adapt computation post hoc, often sacrificing temporal consistency or interpretability. As a result, sparsity is typically treated as an efficiency heuristic rather than an intrinsic property of representation learning.

In this work, we revisit multimodal temporal modeling from an \emph{event-centric} perspective. We define an \emph{event} not as a predefined segment, but as a \textbf{capacity-constrained representation unit} that concentrates task-relevant information within a localized temporal region. Under this view, temporal modeling can be formulated as learning a compact set of such units that optimally allocate representational capacity according to information density.

This perspective naturally leads to a unified framework, \textbf{EvtGraph}, which integrates three components under a shared principle of \emph{budget-constrained information allocation}. First, an event-adaptive compression module reparameterizes dense temporal features into a set of information-concentrated event tokens, allowing salient regions to receive higher effective resolution. Second, a Node Budget Controller enforces a global capacity constraint by selecting a fixed number of tokens, thereby explicitly linking computational cost with representational capacity. Finally, a temporally-constrained sparse graph is constructed over these tokens, imposing a structured prior that restricts dependencies to temporally admissible and local interactions.

Under this formulation, compactness, adaptivity, and sparsity emerge as consequences of a single design principle, rather than independent heuristics. In particular, sparsity arises jointly from (i) information-driven representation compression and (ii) explicit budget constraints, eliminating the need for post hoc pruning.

Our contributions are summarized as follows:
\begin{itemize}[leftmargin=*, noitemsep, topsep=0pt]
    \item We identify a fundamental mismatch between information density and computational allocation in multimodal temporal modeling, and formulate event-centric representation learning as a capacity-constrained problem.
    \item We propose \textsc{EvtGraph}, a unified framework that combines event-adaptive compression, budget-constrained representation learning, and structured temporal graph reasoning.
    \item We show that enforcing an explicit node budget yields a favorable efficiency--accuracy trade-off, achieving strong performance on multimodal clinical prediction (MIMIC-IV) and cross-domain benchmarks (TimeMMD).
    \item We provide detailed analyses demonstrating that the learned representations concentrate on high-information temporal regions and exhibit meaningful structural properties under budget constraints.
\end{itemize}

Taken together, our framework establishes a principled connection between adaptive representation learning and scalable structured reasoning. By treating computation as a constrained resource and aligning it with information density, \textsc{EvtGraph} provides a general approach for modeling high-redundancy multimodal temporal data.

\section{Related Work}
\label{sec:related_work}

\textbf{Event-centric and irregular multimodal temporal modeling.}
Real-world temporal data are often asynchronous and uneven in information density. 
Continuous-time models such as Neural ODEs, Latent ODEs, and Neural CDEs model irregular dynamics in continuous time \citep{chen2018neuralode,rubanova2019latentode,kidger2020neuralcde}, 
while set-based approaches (SeFT, mTAND) handle unaligned observations via permutation-invariant or softly aligned representations \citep{horn2020seft,shukla2021mtand}. 
Multimodal architectures such as Perceiver IO and large vision--language models enable flexible cross-modal fusion \citep{jaegle2021perceiverio,alayrac2022flamingo}. 
However, most approaches still operate on predefined temporal units, which becomes inefficient under highly asynchronous sampling. 
In contrast, our method reparameterizes the temporal axis into event-level tokens, enabling multimodal reasoning over salient events rather than uniformly sampled timesteps.

\textbf{Efficient sequence modeling and adaptive computation.}
A large body of work improves scalability via sparse attention, low-rank approximation, or hardware-aware implementations \citep{kitaev2020reformer,zaheer2020bigbird,dao2022flashattention}. 
Structured state space models (e.g., S4, Mamba) further achieve linear-time sequence modeling through recurrent formulations \citep{gu2022s4,gu2023mamba}. 
While effective for long sequences, these methods still operate over continuous or uniformly discretized time. 
Dynamic token reduction methods (TokenLearner, DynamicViT, ToMe) adapt computation to input content \citep{ryoo2021tokenlearner,rao2021dynamicvit,bolya2023tome}, 
but rely on fixed initial tokenization. 
Our approach differs by jointly learning event-adaptive tokenization and budget-constrained selection, 
transforming dense sequences into compact event-level representations.

\textbf{Information bottleneck and budgeted representations.}
The Information Bottleneck principle characterizes representation learning as a trade-off between compression and task-relevant information \citep{tishby2000information}. 
Our Node Budget Controller can be viewed as an explicit bottleneck that enforces a fixed capacity on event tokens, 
linking predictive utility to computational cost. 
Unlike post-hoc pruning or regularization-based approaches, 
the budget is imposed directly at the representation level, 
enabling efficient learning under strict constraints.

\textbf{Graph compression and temporally constrained reasoning.}
Graph-based methods provide structured modeling for temporal dependencies, 
including graph pooling techniques (DiffPool, SAGPool) and temporal graph networks (TGAT, TGN) \citep{ying2018diffpool,xu2020tgat}. 
Sparse Transformers similarly reduce attention complexity through heuristic sparsity patterns \citep{beltagy2020longformer}. 
However, most approaches either compress an existing dense graph or apply generic sparsification. 
In contrast, T2SG constructs a sparse graph directly over compressed event tokens, 
where edges are constrained by temporal ordering and bounded lag. 
Unlike post-hoc pruning methods that require full graph construction, 
our approach treats node budgeting as a foundational constraint, 
generating efficiency from the outset.

\textbf{Temporal consistency and leakage-free modeling.}
Temporal order is critical for forecasting and clinical prediction, 
as future leakage can lead to overly optimistic results \citep{granger1969causal,runge2020discovering}. 
Rather than performing causal discovery, 
we enforce temporally admissible message passing, 
ensuring that dependencies respect ordering constraints. 
This positions our method as a budget-aware and event-centric framework for temporally consistent multimodal reasoning.

\begin{figure}[t]
\centering
\includegraphics[width=0.95\textwidth]{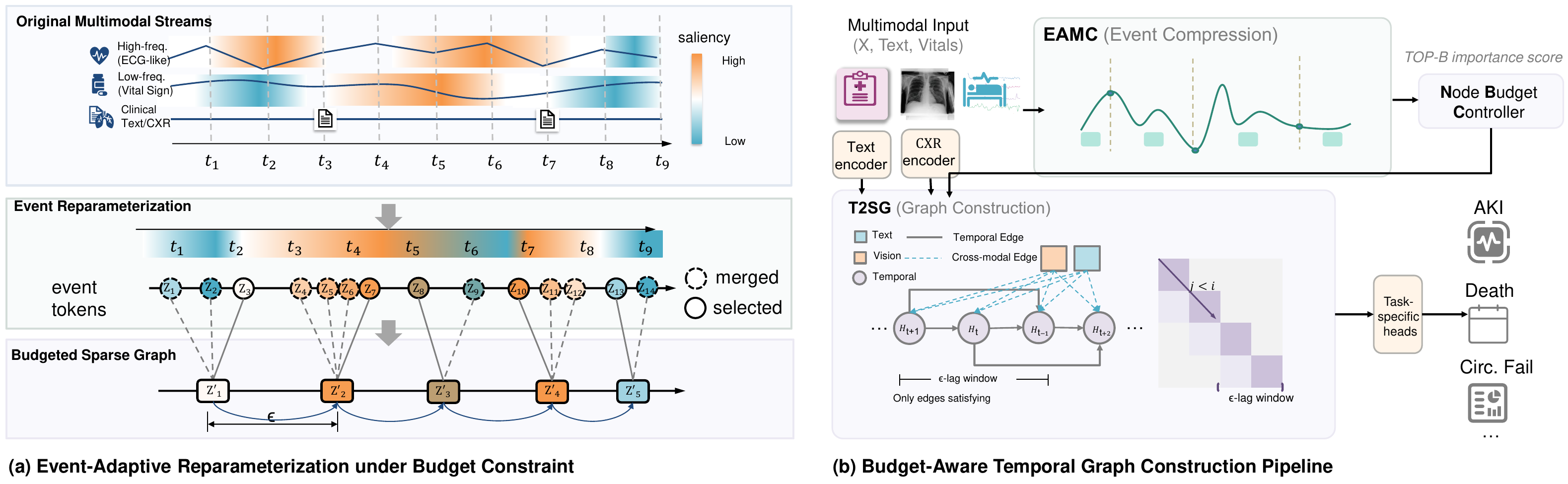}
\caption{
\textbf{Overview of EvtGraph.}
(a) Event-adaptive temporal reparameterization maps multimodal streams into event tokens via soft weights, concentrating representation on salient regions while merging redundant ones.
(b) The selected tokens are organized into a budget-aware temporally constrained graph, where edges respect causal ordering and locality, enabling efficient downstream prediction.
}
\label{fig:overview}
\vspace{-0.8cm}
\end{figure}

\section{Method}
\label{sec:method}

\subsection{Framework Overview}
\label{sec:overview}

We propose \textbf{EvtGraph}, a framework for multimodal temporal learning under explicit computational budgets.
The key idea is to treat representation learning as a \emph{capacity allocation problem}, where limited
representational resources are distributed over time according to information density.
Unlike approaches that adapt computation on top of fixed representations,
EvtGraph formulates representation learning itself under a global capacity constraint.

Given multimodal inputs 
$\mathbf{X} = \{x_t^{(m)}\}_{t=1:T,\, m=1:M}$,
we first encode them into a shared latent space,
yielding temporal features $\mathbf{H} = \{\mathbf{h}_t\}_{t=1}^{T}$.

EvtGraph then performs a progressive transformation:
\[
\mathbf{H}
\;\rightarrow\;
\mathbf{Z}
\;\rightarrow\;
\mathbf{Z}'
\;\rightarrow\;
\mathcal{G},
\]
where dense temporal features are first \emph{reparameterized} into event tokens,
then \emph{compressed under a global budget constraint},
and finally organized into a \emph{temporally constrained sparse graph} for structured reasoning.
In this process, tokenization, selection, and connectivity are jointly determined,
rather than specified independently.

The entire pipeline is trained end-to-end, such that event selection,
budget allocation, and graph structure are jointly optimized for the downstream task.

To handle heterogeneous modalities, all inputs are aligned onto a shared temporal axis:
dense signals are resampled, while sparse modalities are projected to their nearest timestamps.
Missing values are handled via masking rather than imputation to preserve temporal consistency.

This pipeline implements the progressive transformation illustrated in Figure~\ref{fig:overview}(b).

\subsection{Problem Formulation}
\label{sec:problem}

We consider multimodal temporal data $\mathbf{X} = \{x_t^{(m)}\}_{t=1:T,\, m=1:M}$,
and encoded features $\mathbf{H} = \{\mathbf{h}_t\}_{t=1}^{T}$.

A fundamental challenge in temporal modeling is the mismatch between
the non-uniform distribution of information and the uniform allocation of computation.
In real-world signals, informative events are sparse and irregular,
while large portions of the sequence are redundant.

To address this, we seek a compact representation
$\mathbf{Z} = \{\mathbf{z}_s\}_{s=1}^{S}$ with $S \ll T$,
where each token aggregates information across time:
\begin{equation}
\mathbf{z}_s = \sum_{t=1}^{T} w_{s,t} \mathbf{h}_t,
\quad w_{s,t} \ge 0, \quad \sum_t w_{s,t} = 1.
\end{equation}

Rather than viewing Eq.~(1) as a simple weighted aggregation,
we interpret each $\mathbf{z}_s$ as a \emph{capacity-constrained representation unit}
that concentrates task-relevant information within a localized temporal region.
This induces a budgeted representation space, where each token serves as a limited-capacity carrier
of temporal information.
From this perspective, the model implicitly enforces a trade-off between information preservation
and representation compactness, analogous to rate--distortion or information bottleneck principles,
without explicitly optimizing mutual information.
Each token is assigned an importance score $\alpha_s = g_\psi(\mathbf{z}_s)$, and at most $B$ tokens are retained,
so that the budget $B$ directly controls the effective representation capacity and the
compression--performance trade-off.

Over these selected tokens, we define a temporally constrained graph:
\begin{equation}
\mathcal{E} = \{(i,j)\mid 0 < t_i - t_j \le \epsilon\}.
\end{equation}

This formulation jointly constrains both the number of representation units
and the admissible dependencies, linking computational complexity directly
to the structure of the learned representation.

\subsection{Event-Adaptive Representation (EAMC)}
\label{sec:eamc}

We first transform dense temporal features into a compact set of event tokens.
Rather than learning explicit segment boundaries, which is often unstable under weak supervision,
we adopt a fixed coarse partition of the sequence into $S$ contiguous blocks,
each of length $L=\lceil T/S \rceil$.

Within each block, we compute saliency scores
\[
\ell_{s,r} = f_\theta(\mathbf{h}_{s,r}), \quad r=1,\dots,L,
\]
and obtain normalized weights via a temperature-controlled softmax:
\[
w_{s,r} = \frac{\exp(\ell_{s,r}/\tau_e)}{\sum_{r'} \exp(\ell_{s,r'}/\tau_e)}.
\]

The resulting event token is given by
\begin{equation}
\mathbf{z}_s = \sum_{r=1}^{L} w_{s,r}\mathbf{h}_{s,r}.
\end{equation}

Unlike standard attention or pooling mechanisms that operate as feature aggregation operators,
this formulation serves as a \emph{representation reparameterization} of the temporal axis.
Each token $\mathbf{z}_s$ is not merely an attended summary, but a capacity-constrained unit
that redistributes representational mass within a local temporal region.
In particular, the normalization constraint $\sum_r w_{s,r}=1$ induces a competitive allocation
of information within each block, encouraging the model to concentrate on a small subset of
salient timesteps while suppressing redundant ones.

From this perspective, EAMC does not aim to increase expressiveness through richer attention,
but to \emph{reshape the temporal representation space} such that downstream computation
operates on information-dense units. Event structure is therefore not explicitly imposed,
but emerges implicitly from the learned distribution of information density.

To stabilize training, we gradually anneal $\tau_e$ and introduce a residual term:
\begin{equation}
\mathbf{z}_s =
(1-\gamma)\sum_{r} w_{s,r}\mathbf{h}_{s,r}
+ \gamma \cdot \mathrm{Pool}(\{\mathbf{h}_{s,r}\}),
\end{equation}
which prevents degeneration under overly sharp distributions.

\subsection{Node Budget Allocation}
\label{sec:nbc}

Given event tokens $\mathbf{Z}$, we retain at most $B$ tokens under a global budget constraint.
Each token is assigned an importance score $\alpha_s = g_\psi(\mathbf{z}_s)$,
and the top-$B$ tokens are selected.

From a representation perspective, the node budget $B$ imposes an explicit \emph{capacity constraint}
on the latent space, forcing the model to allocate a limited number of representation slots
to the most informative temporal regions.
Unlike conventional token pruning, where selection is applied as a post-hoc efficiency heuristic,
the budget here acts as a \emph{first-class modeling constraint},
directly shaping the learned representation and its effective capacity.

To enable gradient-based optimization, we introduce a differentiable relaxation.
For each selected token $\hat{\mathbf{z}}_i$, we define a soft aggregation:
\begin{equation}
\tilde{\mathbf{z}}_i = \sum_{s} q_{i,s}\mathbf{z}_s,
\end{equation}
where
\[
q_{i,s} \propto \exp\!\left((\alpha_s - \lambda |t_s - t_i|)/\tau_B\right).
\]

Rather than merely approximating top-$B$ selection, this relaxation performs
\emph{localized information redistribution}: tokens compete globally via $\alpha_s$,
while the temporal bias $|t_s - t_i|$ encourages nearby information to be reassigned
instead of discarded.

A straight-through estimator is used so that forward computation follows hard selection,
while gradients are propagated through the soft aggregation.

To mitigate information loss from hard selection, discarded tokens are further aggregated
to their nearest selected token in time, yielding
\begin{equation}
\mathbf{z}'_i = \mathrm{LayerNorm}\!\left(
\hat{\mathbf{z}}_i + \frac{1}{|\mathcal{R}_i|+\delta}
\sum_{s\in\mathcal{R}_i} \mathbf{z}_s
\right).
\end{equation}

This results in a compact set of nodes $\mathbf{Z}'$ that preserves local context
while respecting the global budget constraint.

\subsection{Temporally-Constrained Sparse Graph (T2SG)}
\label{sec:t2sg}

We construct a sparse temporal graph over the selected tokens $\mathbf{Z}'$.
Edges are restricted by temporal admissibility:
\[
(i,j) \in \mathcal{E}
\iff
j < i \ \wedge\ t_i - t_j \le \epsilon,
\]
ensuring causal and leakage-free message passing.

Rather than introducing sparsity through generic pruning or attention masking,
this construction derives the graph structure directly from the \emph{compressed event representation}.
As a result, the sparsity pattern is not imposed independently, but is coupled with
both the node budget and the temporal distribution of information.

This constraint can be interpreted as a structured prior over temporal dependencies,
encoding the inductive bias that relevant interactions are both temporally ordered
and locally bounded.

To further control complexity, each node only retains its top-$\kappa$ neighbors,
leading to the update
\begin{equation}
\mathbf{z}_i^{(l+1)} =
\sigma\!\left(
\sum_{j \in \mathcal{N}_\kappa(i)}
\alpha_{ij}^{(l)} \mathbf{W}^{(l)} \mathbf{z}_j^{(l)}
\right).
\end{equation}

The resulting graph contains at most $B$ nodes and
$O(\kappa B)$ edges, yielding near-linear complexity.
Importantly, sparsity emerges jointly from (i) event-level compression and
(ii) explicit budget constraints, rather than from post-hoc edge pruning.

\subsection{Implementation Details}
\label{sec:method_detail}

Graph sparsity is controlled by the temporal window $\epsilon$ and neighbor budget $\kappa$, 
which enforce locality while reducing redundant connections.
Given the small node budget $B$, we use shallow graph layers ($L=2$--$3$) to avoid over-smoothing.

Node features are normalized with LayerNorm to stabilize cross-modal similarity, 
while temporal consistency is ensured via strict admissibility constraints.

Ablations confirm that removing event adaptivity, budget control, or temporal constraints degrades performance or increases complexity, 
highlighting the importance of their joint design.

\section{Experiments}
\label{sec:experiments}

We evaluate the proposed framework from three complementary perspectives:
(1) predictive performance across multimodal tasks,
(2) efficiency--information trade-offs under explicit budget constraints,
and (3) structural validity of the learned event representations and temporal dependencies.

Concretely, we examine whether the model (i) improves accuracy under constrained computation,
(ii) preserves information through event-adaptive compression and budgeted selection,
and (iii) learns meaningful temporal structures aligned with data semantics.
Experiments are conducted on clinical prediction, cross-domain forecasting, and sensor-based activity recognition benchmarks.

\subsection{Experimental Setup}
\label{sec:experiments_setup}

\textbf{Datasets.}~~
We evaluate on three complementary benchmarks covering clinical prediction and cross-domain generalization:  
(1) \textbf{MIMIC-IV + CXR}, a multimodal clinical dataset with irregular sampling and missing observations; 
(2) \textbf{TimeMMD}, a cross-domain multimodal benchmark with paired temporal and textual data.
(3) \textbf{UCI HAR}, a lightweight multivariate sensor dataset for human activity recognition, used to evaluate generalization to short event-like temporal sequences.

\textbf{Tasks and Metrics.}~~
We consider regression, classification, and efficiency evaluation. 
Regression tasks include physiological prediction on MIMIC and forecasting on TimeMMD (MSE, MAE, RMSE). 
Classification tasks involve clinical risk prediction, evaluated using AUROC and AUPRC. 
Efficiency is measured in terms of latency, memory, and computational cost under varying sequence lengths and node budgets.

\textbf{Baselines.}~~
We compare against representative methods across four paradigms: 
(1) \textbf{Discrete-time models} (GRU-D, LSTM, TimesNet, iTransformer, PatchTST), 
(2) \textbf{Continuous-time models} (Neural CDE, Latent ODE, mTAN, SeFT), 
(3) \textbf{Adaptive computation methods} (ToMe, DynamicViT), 
and (4) \textbf{Graph-based models} (Perceiver-IO, TGN, DiffPool, MinCutPool). 
This set covers key design axes including temporal dynamics, adaptive computation, and structured reasoning. For forecasting-oriented baselines, we tune sequence aggregation strategies(mean pooling, last-token pooling, attention pooling) on the validation set and
report the best configuration.

\textbf{Implementation Details.}~~
All models are trained under a unified setting for fair comparison. 
Additional details on preprocessing, optimization, and hyperparameters are provided in Appendix.\ref{app:exp_setup}.

\begin{table}[t]
\centering
\caption{Main results on MIMIC-IV + CXR. Results are AUROC over 3 seeds. Best results are in \textbf{bold}, and second best are \underline{underlined}.}
\label{tab:mimic_main}
\scriptsize
\setlength{\tabcolsep}{2.5pt}
\renewcommand{\arraystretch}{1.05}

\begin{adjustbox}{width=\textwidth}
\begin{tabular}{l l c c c c c c}
\toprule
\textbf{Category} & \textbf{Model} & \textbf{Params} 
& \textbf{Macro} & \textbf{AKI} & \textbf{Circ. Fail} 
& \textbf{Death} & \textbf{Sepsis} \\
\midrule

\multirow{3}{*}{\textbf{Continuous-Time}}
& Neural CDE     
& 1.6M  
& 0.5578$\pm$0.011 
& 0.5182$\pm$0.012 
& 0.5235$\pm$0.008 
& 0.4992$\pm$0.005 
& 0.6904$\pm$0.037 \\

& Latent ODE     
& 121K  
& 0.7971$\pm$0.005 
& 0.7374$\pm$0.015 
& 0.8621$\pm$0.006 
& 0.7248$\pm$0.014 
& 0.8641$\pm$0.023 \\

& mTAN           
& 60K   
& 0.7328$\pm$0.005 
& 0.6933$\pm$0.009 
& 0.7084$\pm$0.009 
& 0.7344$\pm$0.013 
& 0.7950$\pm$0.011 \\

\midrule
\multirow{2}{*}{\textbf{Adaptive Computation}}
& ToMe           
& 180K  
& 0.7974$\pm$0.015 
& 0.7274$\pm$0.014 
& 0.8026$\pm$0.019 
& 0.7949$\pm$0.016 
& 0.8648$\pm$0.017 \\

& DynamicViT     
& 183K  
& 0.7916$\pm$0.006 
& 0.7209$\pm$0.006 
& 0.7936$\pm$0.013 
& 0.7798$\pm$0.006 
& 0.8722$\pm$0.012 \\

\midrule
\multirow{2}{*}{\textbf{Graph Pooling}}
& DiffPool       
& 25K   
& 0.4919$\pm$0.017 
& 0.5116$\pm$0.020 
& 0.5177$\pm$0.017 
& 0.4780$\pm$0.021 
& 0.4604$\pm$0.076 \\

& MinCutPool     
& 25K   
& 0.5807$\pm$0.019 
& 0.5778$\pm$0.014 
& 0.5833$\pm$0.010 
& 0.5436$\pm$0.063 
& 0.6181$\pm$0.073 \\

\midrule
\multirow{5}{*}{\textbf{Discrete-Time TS}}
& LSTM           
& 102K  
& \underline{0.8405$\pm$0.006} 
& \underline{0.7981$\pm$0.006} 
& \underline{0.8704$\pm$0.001} 
& 0.8089$\pm$0.011 
& 0.8848$\pm$0.019 \\

& GRU            
& 77K   
& 0.8392$\pm$0.001 
& 0.7887$\pm$0.002 
& 0.8566$\pm$0.008 
& \underline{0.8224$\pm$0.015} 
& \underline{0.8889$\pm$0.024} \\

& TCN            
& 77K   
& 0.8127$\pm$0.017 
& 0.7504$\pm$0.009 
& 0.8419$\pm$0.007 
& 0.7971$\pm$0.020 
& 0.8615$\pm$0.041 \\

& Transformer    
& 130K  
& 0.8006$\pm$0.015 
& 0.7327$\pm$0.005 
& 0.8101$\pm$0.010 
& 0.7951$\pm$0.014 
& 0.8643$\pm$0.034 \\

& TimesNet       
& 18.9M 
& 0.8002$\pm$0.003 
& 0.7349$\pm$0.006 
& 0.7934$\pm$0.008 
& 0.8148$\pm$0.004 
& 0.8578$\pm$0.027 \\

\midrule
\multirow{2}{*}{\textbf{TS-SOTA}}
& iTransformer   
& 137K  
& 0.7795$\pm$0.006 
& 0.6992$\pm$0.006 
& 0.7642$\pm$0.009 
& 0.8141$\pm$0.009 
& 0.8407$\pm$0.015 \\

& PatchTST       
& 342K  
& 0.7744$\pm$0.011 
& 0.7026$\pm$0.005 
& 0.7509$\pm$0.010 
& 0.8209$\pm$0.012 
& 0.8232$\pm$0.024 \\

\midrule
\textbf{Ours}
& \textbf{EvtGraph} 
& \textbf{286K} 
& \textbf{0.9060$\pm$0.007} 
& \textbf{0.9043$\pm$0.005} 
& \textbf{0.9037$\pm$0.006} 
& \textbf{0.9147$\pm$0.024} 
& \textbf{0.8961$\pm$0.003} \\

\bottomrule
\end{tabular}
\end{adjustbox}
\vspace{-0.5mm}
\end{table}

\subsection{Multimodal Clinical and Cross-Domain Performance}
\label{sec:main_results}

EvtGraph achieves consistent improvements on multimodal clinical prediction tasks (Table~\ref{tab:mimic_main}), 
indicating robust gains across all outcomes.

We attribute this to its ability to filter redundant temporal regions and concentrate computation on salient events, 
which is particularly beneficial under irregular and noisy clinical signals where global attention may be less effective.

\begin{figure*}[t]
    \centering
    \includegraphics[width=0.98\textwidth]{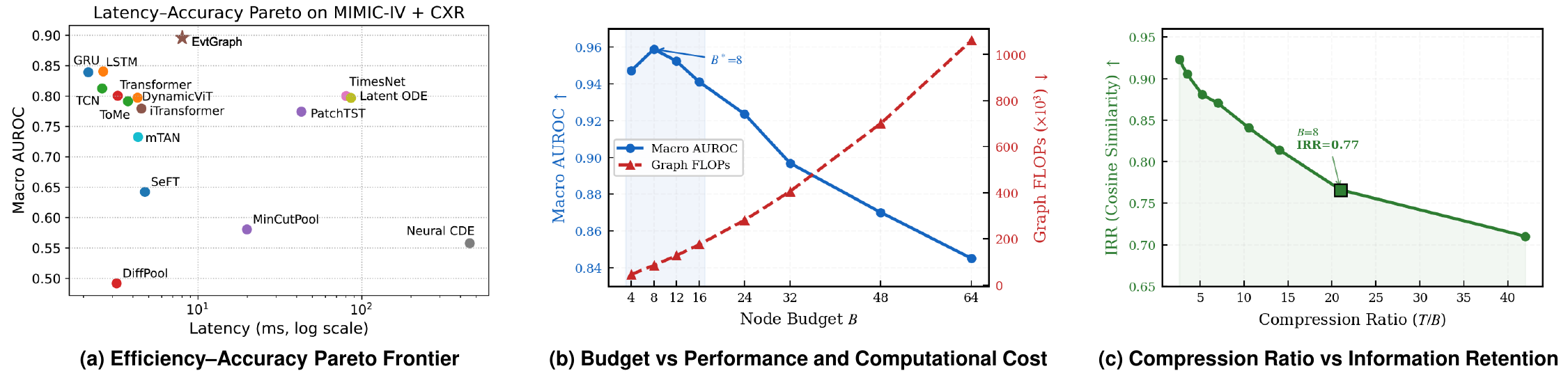}
    \caption{
\textbf{Efficiency--information trade-off under budget-aware modeling.}
(a) Efficiency--accuracy Pareto frontier on MIMIC-IV + CXR, showing strong performance under reduced latency.
(b) Performance and computational cost versus node budget $B$, with AUROC peaking at a small budget ($B^*=8$).
(c) Compression ratio versus information retention, where IRR remains high under aggressive compression.
}
    \label{fig:efficiency_tradeoff}
\end{figure*}

As shown in Figure~\ref{fig:efficiency_tradeoff}(a), EvtGraph further exhibits a favorable efficiency--accuracy trade-off, 
achieving the best Macro AUROC under comparable or lower computational cost.

On the TimeMMD benchmark (Table~\ref{tab:timemmd_main}), EvtGraph attains the lowest cross-domain error, 
while maintaining competitive in-domain performance, suggesting improved robustness under distribution shift.
Consistent gains are also observed on UCI HAR, indicating that the benefits extend to shorter and more structured temporal sequences 
(Appendix~\ref{app:har}).

\begin{table}[t]
\centering
\caption{
\textbf{Results on TimeMMD and UCI HAR.}
For TimeMMD, we report MSE and MAE (mean $\pm$ std), where lower is better.
For UCI HAR, we report accuracy, where higher is better.
Best results are in \textbf{bold}, and second best are \underline{underlined}.
}
\label{tab:timemmd_main}

\scriptsize
\setlength{\tabcolsep}{2.2pt}
\renewcommand{\arraystretch}{1.05}

\begin{tabular*}{\textwidth}{@{\extracolsep{\fill}} l l c c c c c c}
\toprule
\textbf{Category} & \textbf{Model} & \textbf{Params}
& \multicolumn{2}{c}{\textbf{In-domain}} 
& \multicolumn{2}{c}{\textbf{Cross-domain}}
& \textbf{UCI HAR} \\
\cmidrule(lr){4-5} \cmidrule(lr){6-7}
& & 
& \textbf{MSE} & \textbf{MAE}
& \textbf{MSE} & \textbf{MAE}
& \textbf{Acc.} \\
\midrule

\multirow{2}{*}{\textbf{RNN}}
& LSTM 
& 337K 
& 0.226$\pm$0.005 & 0.385$\pm$0.006
& 1.973$\pm$0.160 & 1.125$\pm$0.046
& 0.9230 \\

& GRU  
& 287K 
& \underline{0.217$\pm$0.007} & \underline{0.368$\pm$0.006}
& 2.056$\pm$0.264 & 1.159$\pm$0.076
& 0.9240 \\

\midrule
\multirow{1}{*}{\textbf{Conv}}
& TCN  
& 287K 
& 0.621$\pm$0.202 & 0.580$\pm$0.097
& 2.459$\pm$1.041 & 1.178$\pm$0.264
& \underline{0.9430} \\

\midrule
\multirow{1}{*}{\textbf{Attention}}
& Transformer 
& 565K 
& 0.251$\pm$0.032 & 0.372$\pm$0.026
& 1.430$\pm$0.034 & 0.957$\pm$0.006
& 0.8870 \\

\midrule
\multirow{2}{*}{\textbf{Continuous-Time}}
& Latent ODE 
& 262K 
& 0.237$\pm$0.035 & 0.353$\pm$0.032
& 1.595$\pm$0.141 & 1.010$\pm$0.046
& 0.8867 \\

& mTAN       
& 241K 
& 0.279$\pm$0.009 & 0.395$\pm$0.007
& 2.115$\pm$0.399 & 1.182$\pm$0.109
& 0.8894 \\

\midrule
\multirow{2}{*}{\textbf{Adaptive Computation}}
& DynamicViT 
& 776K 
& 0.246$\pm$0.043 & 0.375$\pm$0.042
& \underline{1.285$\pm$0.140} & \underline{0.902$\pm$0.062}
& 0.8928 \\

& ToMe       
& 763K 
& 0.227$\pm$0.007 & 0.361$\pm$0.001
& 1.385$\pm$0.198 & 0.940$\pm$0.060
& 0.8992 \\

\midrule
\multirow{2}{*}{\textbf{Graph Pooling}}
& MinCutPool 
& 170K 
& 0.438$\pm$0.117 & 0.489$\pm$0.067
& 1.738$\pm$0.262 & 1.028$\pm$0.084
& 0.9074 \\

& DiffPool   
& 170K 
& 0.451$\pm$0.115 & 0.499$\pm$0.066
& 1.756$\pm$0.274 & 1.033$\pm$0.086
& 0.8476 \\

\midrule
\textbf{Ours}
& \textbf{EvtGraph}
& \textbf{149K}
& \textbf{0.203$\pm$0.025} & \textbf{0.337$\pm$0.015}
& \textbf{1.054$\pm$0.161} & \textbf{0.812$\pm$0.056}
& \textbf{0.9494} \\

\bottomrule
\end{tabular*}
\vspace{-0.5mm}
\end{table}
% =========================================================
% Fig. 4: Efficiency--information trade-off
% =========================================================

\subsection{Efficiency--Information Trade-off and Event-Adaptive Representation}
\label{sec:efficiency_event}

We evaluate the trade-off between predictive performance, computational cost,
and information retention under varying node budgets $B$.

As shown in Figure~\ref{fig:efficiency_tradeoff}, EvtGraph achieves a favorable
efficiency--accuracy Pareto frontier, and performance peaks at a small budget
($B^* = 8$) before saturating as computational cost increases. This suggests that only a limited number of event tokens is sufficient to capture
the dominant temporal structure.

\begin{figure*}[t]
    \centering
    \includegraphics[width=0.98\textwidth]{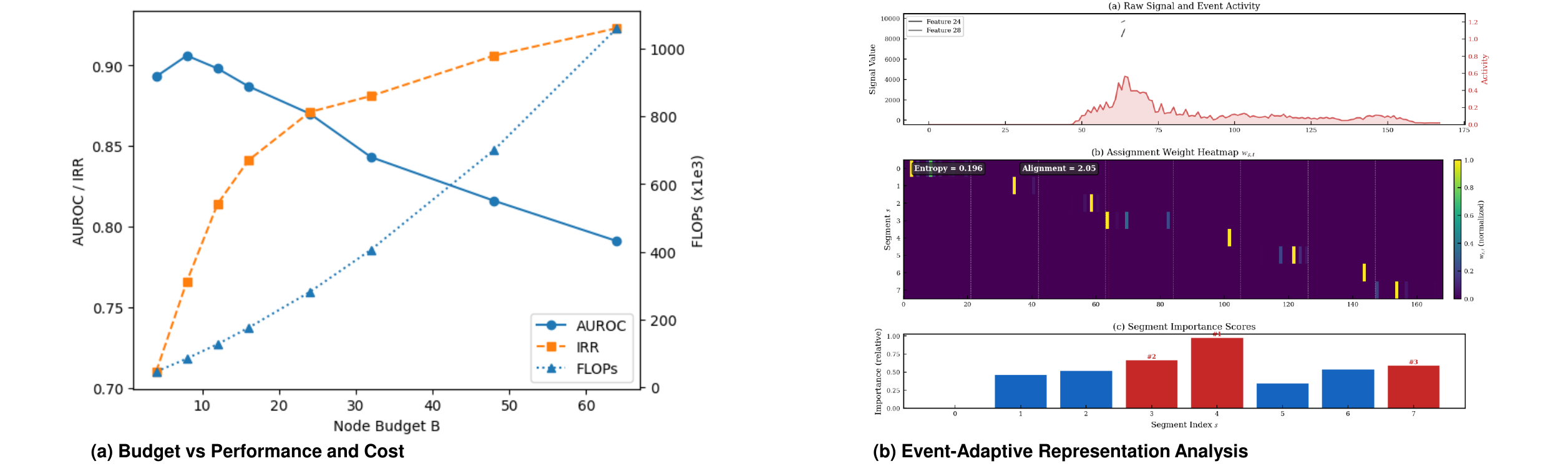}
    \caption{
\textbf{Event-adaptive representation analysis.}
(a) Raw signal and activity patterns highlighting salient temporal regions.
(b) Learned event-adaptive representations. The assignment heatmap $w_{s,t}$ shows highly concentrated temporal allocation, while segment importance scores $\alpha_s$ indicate that only a few event tokens dominate the representation.
}
    \label{fig:event_representation}
    \vspace{-0.4mm}
\end{figure*}

To examine whether this efficiency arises from meaningful representations rather than
uniform compression, we analyze the learned event-adaptive structure in
Figure~\ref{fig:event_representation}. As shown in Figure~\ref{fig:event_representation}(b), the assignment weights $w_{s,t}$ concentrate on a small subset of salient temporal locations, while the segment
importance scores $\alpha_s$ indicate selective allocation of representational capacity.
Over 94\% of the total weight is concentrated within the top-3 timesteps per segment,
demonstrating highly selective computation. Together, these results suggest that EvtGraph achieves efficiency not through uniform
compression, but by allocating computation to high-information temporal regions.

% =========================================================
% Fig. 5: Event-adaptive representation analysis
% =========================================================

% =========================================================
% Fig. 6: Graph diagnostics
% =========================================================
\begin{figure*}[t]
    \centering
    \includegraphics[width=0.98\textwidth]{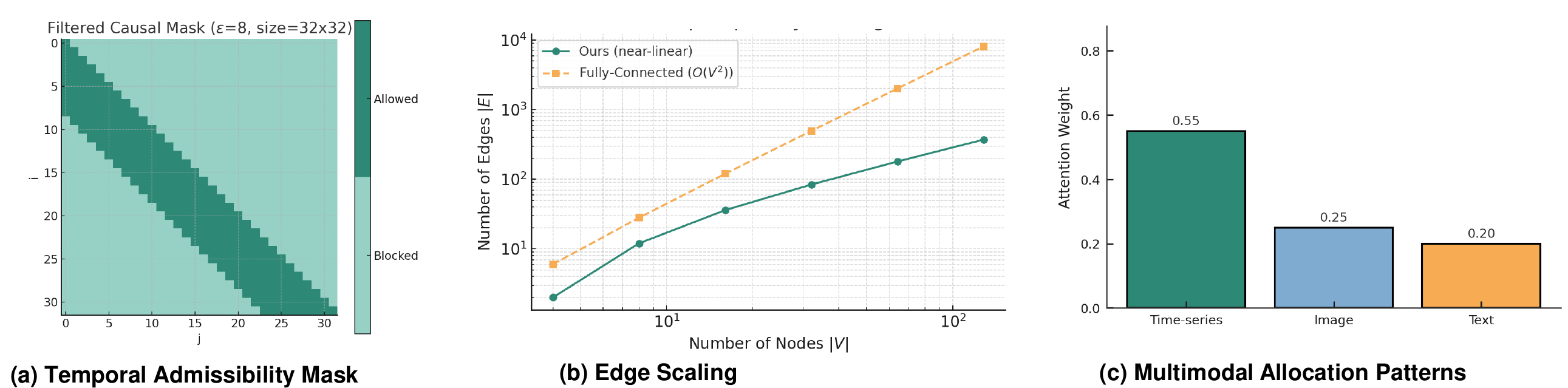}
    \caption{
\textbf{Structural diagnostics of the budget-aware temporal graph.}
(a) Temporal admissibility mask under lag constraint $\epsilon$, 
showing that edges are restricted to past nodes within a bounded window.
(b) Edge count as a function of node count, demonstrating near-linear scaling 
($\mathcal{O}(\kappa B)$) compared to quadratic growth in fully connected graphs.
(c) Modality-wise allocation weights, indicating adaptive capacity distribution 
across time-series, image, and text modalities.
}
    \label{fig:graph_diagnostics}
\end{figure*}

\subsection{Budget and Graph Diagnostics}
\label{sec:graph_analysis}

Figure~\ref{fig:graph_diagnostics}(a) visualizes the temporally admissible mask induced
by the lag constraint $\epsilon$. Figure~\ref{fig:graph_diagnostics}(b) shows that the
number of edges scales near-linearly with the number of retained nodes, consistent with
$\mathcal{O}(\kappa B)$. Figure~\ref{fig:graph_diagnostics}(c) further illustrates
modality-wise allocation, showing that the model adaptively distributes capacity across
time-series, image, and text modalities.

\subsection{Ablation Study}
\label{sec:ablation}

\begin{wraptable}{r}{0.58\linewidth}
\vspace{-6pt}
\centering
\small
\setlength{\tabcolsep}{3pt}
\renewcommand{\arraystretch}{1.05}

\caption{
\textbf{Ablation on MIMIC-IV + CXR.}
}
\label{tab:ablation}

\begin{tabular}{lccccc}
\toprule
Variant 
& Macro 
& AKI 
& Circ. Fail
& Death 
& Sepsis \\
\midrule
Full
& \textbf{0.906}
& \textbf{0.904}
& \textbf{0.9037}
& 0.915
& \textbf{0.896} \\
\midrule
w/o NBC 
& 0.898 & 0.891 & 0.880 & \textbf{0.919} & 0.880 \\
w/o Temp 
& 0.895 & 0.890 & 0.8446 & 0.908 & 0.882 \\
w/o Event 
& 0.879 & 0.867 & 0.898 & 0.895 & 0.855 \\
w/o RevIN 
& 0.835 & 0.775 & 0.884 & 0.903 & 0.796 \\
w/o Mask 
& 0.867 & 0.838 & 0.899 & 0.914 & 0.835 \\
\midrule
Uniform Compression 
& 0.894 & 0.892 & 0.904 & 0.907 & 0.887 \\
Random Selection 
& 0.892 & 0.890 & 0.900 & 0.899 & 0.882 \\
\bottomrule
\end{tabular}

\end{wraptable}

We conduct ablations on MIMIC-IV + CXR (Table~\ref{tab:ablation}).
Removing event adaptivity leads to the largest drop, highlighting its central role in allocating capacity to informative temporal regions.
Performance also degrades without the budget controller, suggesting its regularization effect, while \texttt{w/o Temp} underperforms, confirming the necessity of temporally admissible graph construction.
Removing RevIN and masking further reduces performance, indicating the importance of distribution alignment and leakage-free modeling.

To isolate the effect of adaptivity, we compare against two controlled baselines under the same node budget:
\textit{Uniform Compression} and \textit{Random Selection}.
Both underperform the full model, indicating that improvements arise not merely from compression, but from adaptive concentration of computation on salient regions.

Overall, gains stem from the joint effect of event-adaptive representation, budget control, and temporally constrained reasoning.
\section{Conclusion}

We present \textbf{EvtGraph}, a framework for multimodal temporal modeling under explicit computational constraints.
By casting representation learning as a \emph{capacity allocation problem}, EvtGraph adaptively distributes a fixed budget over time according to information density, producing a compact set of event-level tokens for structured reasoning via a temporally constrained graph.

Under this formulation, compression, selection, and sparsity are not introduced as independent heuristics, but emerge jointly from a unified budget-aware design.
Empirically, EvtGraph achieves strong performance across multimodal clinical and cross-domain benchmarks, while consistently improving efficiency--accuracy trade-offs under limited computation.

\paragraph{Limitations.}
EvtGraph relies on coarse temporal partition, which may limit ability to capture events with highly variable or long-range structure. Its advantages may diminish for near-stationary signals, and performance depends on the choice of node budget $B$, which requires task-specific tuning.

\paragraph{Broader Impact.}
EvtGraph provides a practical approach for modeling high-redundancy temporal data in resource-constrained settings.
More broadly, it suggests a perspective in which representation learning is governed by explicit capacity constraints, offering a principled pathway toward scalable and efficient modeling in structured domains.

\bibliographystyle{plainnat}  % 推荐
\bibliography{references} 
%%%%%%%%%%%%%%%%%%%%%%%%%%%%%%%%%%%%%%%%%%%%%%%%%%%%%%%%%%%%

\appendix

\section{Broader Impact}

This work aims to improve the efficiency and interpretability of multimodal machine learning, 
with a primary focus on healthcare applications. 
The proposed framework enables modeling long-term patient records under strict computational budgets, 
potentially improving accessibility of clinical decision support systems in resource-constrained settings.

However, models trained on historical clinical data (e.g., MIMIC-IV) may inherit biases related to 
demographics, data collection processes, or clinical practices. 
While our method does not explicitly address bias, its emphasis on event-adaptive representations 
and structured temporal modeling may facilitate auditing of which temporal segments and modalities 
contribute to predictions, supporting more transparent analysis.

Although no immediate harmful applications are identified, risks such as bias amplification or misuse 
in automated decision-making remain possible. 
Therefore, we emphasize that the proposed method should be deployed with appropriate human oversight 
and in conjunction with domain expertise.

\section{Code Availability}

An anonymized implementation is provided at:
\href{https://anonymous.4open.science/status/EvtGraph-F279}{anonymous.4open.science (EvtGraph-F279)}.
\section{Notation}
\label{app:notation}

\begin{table}[htbp]
\centering
\caption{Notation used in the proposed framework.}
\label{tab:notation}
\begin{tabular}{llc}
\toprule
\textbf{Symbol} & \textbf{Description} & \textbf{Shape / Domain} \\
\midrule

$X=\{x_t\}_{t=1}^T$ 
& Multimodal time-series inputs 
& $T \times d_0$ \\

$H=\{h_t\}_{t=1}^T$ 
& Encoded temporal features 
& $T \times d$ \\

$Z=\{z_s\}_{s=1}^S$ 
& Event-level tokens 
& $S \times d$ \\

$w_{s,t}$ 
& Assignment weights (time $\to$ event tokens) 
& $[0,1],\ \sum_t w_{s,t}=1$ \\

$\alpha_s$ 
& Event importance score 
& $\mathbb{R}$ \\

$\mathcal{S}_B$ 
& Index set of Top-$B$ selected tokens 
& $|\mathcal{S}_B| \le B$ \\

$\tilde{z}_i$ 
& Soft aggregated token (training) 
& $\mathbb{R}^d$ \\

$z'_i$ 
& Final retained token after aggregation 
& $\mathbb{R}^d$ \\

\midrule

$\mathcal{G}=(\mathcal{V},\mathcal{E})$ 
& Temporal graph over selected tokens 
& -- \\

$\mathcal{V}$ 
& Node set ($=\{z'_i\}$) 
& $\hat S$ nodes \\

$\mathcal{E}$ 
& Edge set under temporal constraints 
& -- \\

$M[i,j]$ 
& Temporal admissibility mask 
& $\{0,1\}$ \\

$\epsilon$ 
& Temporal lag window 
& $\mathbb{N}$ \\

$\kappa$ 
& Max neighbors per node 
& $\mathbb{N}$ \\

$h^{(\ell)}$ 
& Node representations at layer $\ell$ 
& $\hat S \times d$ \\

\midrule

$B$ 
& Node budget 
& $\mathbb{N}$ \\

$\hat S$ 
& Number of retained nodes 
& $\le B$ \\

$\tau_B$ 
& Temperature for Top-$B$ relaxation 
& $\mathbb{R}_{>0}$ \\

\bottomrule
\end{tabular}
\end{table}
\section{Experimental Setup}
\label{app:exp_setup}

\subsection{MIMIC-IV + CXR Preprocessing and Protocol}

We construct a multimodal cohort by aligning EHR data from 
MIMIC-IV \cite{johnson2023mimiciv} 
and chest X-ray data from MIMIC-CXR \cite{johnson2019mimiccxr}.
For each patient encounter, we define a fixed observation window of 7 days after admission, corresponding to 168 hours.
All dynamic variables are discretized at a 2-hour resolution, yielding $T=84$ time steps.
The time-series branch uses $F=20$ dynamic clinical variables, while static covariates such as demographic or admission-level attributes are processed separately when used by multimodal baselines.

Time-series inputs are represented as tensors $\mathbf{X}\in\mathbb{R}^{B\times T\times F}$ together with explicit missing-value masks.
We do not impute missing values; instead, missingness is provided as part of the input representation.
Continuous variables are normalized using statistics computed only from the training split.
CXR images are resized to $224\times224$ and normalized using standard ImageNet statistics.
Radiology reports are encoded using a pretrained text encoder and aligned to the timestamp of the corresponding CXR examination.
Sparse modalities are assigned to their nearest valid timestamp within the observation window.

We consider four binary clinical prediction tasks: acute kidney injury (AKI), circulatory failure, sepsis, and death.
For each task, the input consists only of measurements, images, and reports observed within the 7-day window.
Labels are defined from clinical events occurring after the observation window, so that the model predicts future outcomes rather than reconstructing already observed events.
Unless otherwise specified, outcomes that occur within the observation window are excluded from the positive target definition to avoid label leakage.
Thus, each task is formulated as
\[
    \hat{y} = f(\mathcal{X}_{[0,T_{\mathrm{obs}}]}),
    \qquad
    y = \mathbb{I}\{\text{target event occurs after } T_{\mathrm{obs}}\},
\]
where $T_{\mathrm{obs}}$ denotes the end of the observation window.

To avoid temporal leakage from multimodal inputs, we enforce strict timestamp filtering.
CXR images and radiology reports are included only if their acquisition or report timestamps fall within the observation window.
Reports generated after the observation window are removed, even if they correspond to earlier studies.
When multiple CXR studies are available, each image--report pair is aligned using its own timestamp, and future studies are never exposed to the model.
This prevents textual descriptions or diagnostic impressions from revealing outcomes that occur after the prediction time.

We remove variables with extremely high missing rates or low support and apply per-feature normalization using training-set statistics.
The final cohort contains 5,506 patient encounters, split at the patient level into training, validation, and test sets with ratio $0.7/0.15/0.15$.
All visits from the same patient are assigned to the same split to prevent patient-level information leakage.

All baselines are evaluated under the same preprocessing pipeline, patient-level splits, observation windows, and target definitions.
For multimodal baselines, the same time-series variables, missing-value masks, static covariates, CXR images, and radiology reports are provided whenever the architecture supports the corresponding modality.
When a baseline is originally designed for unimodal time-series modeling, we use the same encoded modality features and fuse them with the temporal representation through a shared late-fusion MLP, so that performance differences are not caused by access to different input information.
For forecasting-oriented architectures, we replace the forecasting head with the same binary classification head and tune temporal aggregation strategies on the validation set.

\subsection{Time-MMD Dataset and Preprocessing.} 
\begin{figure}[t]
    \centering
    \includegraphics[width=\linewidth]{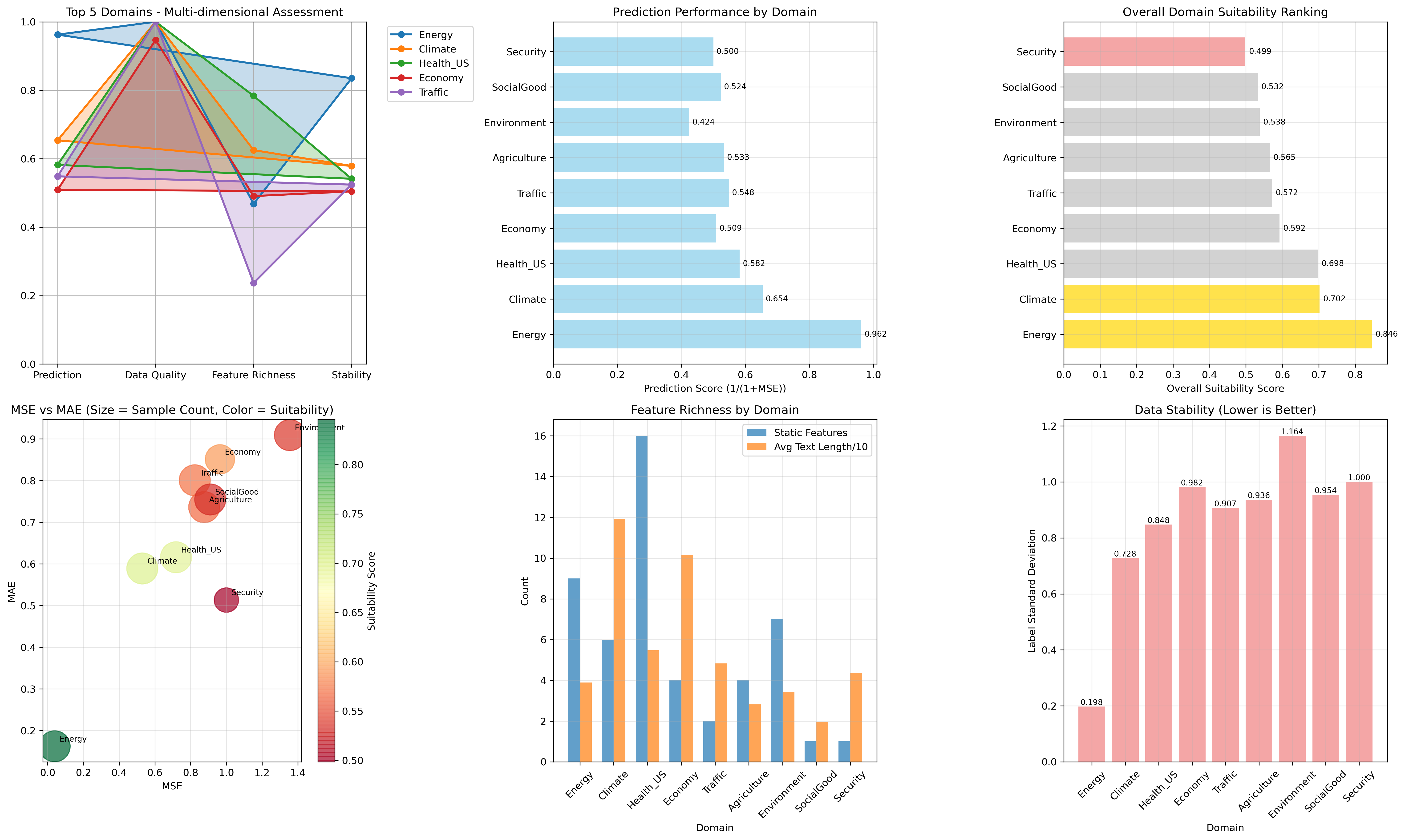}
    \caption{Domain-level suitability analysis of Time-MMD. 
    Multi-dimensional evaluation includes prediction performance, feature richness, stability, 
    and overall ranking across nine domains. Energy and Climate emerge as top candidates.}
    \label{fig:domain_analysis}
\end{figure}

\begin{figure}[htbp]
    \centering
    \includegraphics[width=\linewidth]{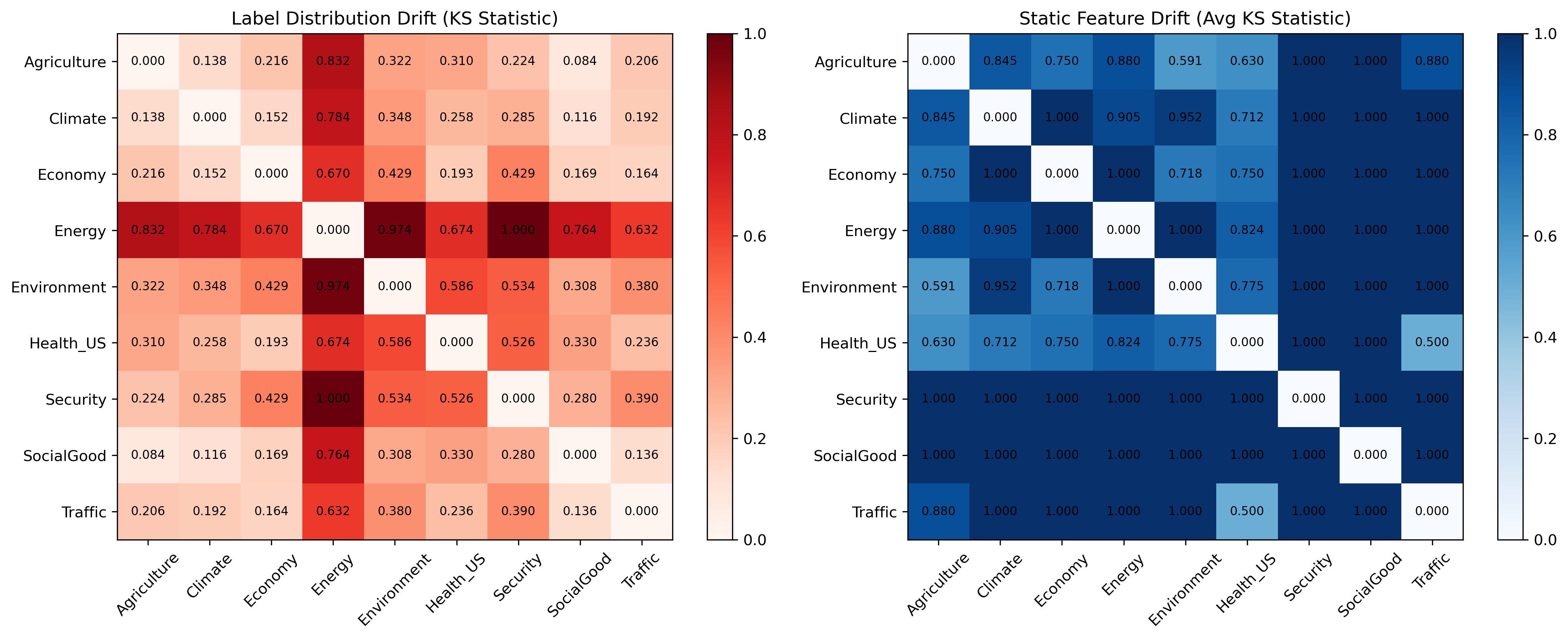}
    \caption{Cross-domain drift analysis of Time-MMD. 
    Left: label distribution drift measured by KS statistic. 
    Right: static feature drift averaged across features. 
    Larger values indicate greater divergence between domains, highlighting challenges of generalization.}
    \label{fig:drift_analysis}
\end{figure}

We also conduct experiments on the Time-MMD dataset, which integrates nine domains (Energy, Climate, Health\_US, Economy, Traffic, Agriculture, Environment, SocialGood, Security) with three modalities: time-series signals, free-text records, and structured static features. 
Each domain contains $309$--$500$ samples for 24-hour forecasting tasks. 
Time-series are transformed via first-order differencing 
$\Delta y_t = y_t - y_{t-1}$ 
followed by normalization (z-score or robust scaling). 
Static features are standardized independently, while text records are aligned by nearest timestamp. 
Domain-level statistics (prediction performance, feature richness, stability) are visualized in Figure~\ref{fig:domain_analysis}, 
and cross-domain drift is illustrated in Figure~\ref{fig:drift_analysis}. 
The analysis reveals that Energy and Climate domains are the most suitable: 
Energy provides highly stable numerical signals with lowest error (MSE = 0.039, MAE = 0.162), 
while Climate offers the richest textual information (average length $119.3$ characters) with strong seasonality patterns. 
Thus, these two domains are selected as primary benchmarks for multimodal forecasting. 

\subsection{UCI HAR Dataset and Preprocessing.}

We further evaluate the proposed framework on the UCI Human Activity Recognition (HAR) dataset,
a lightweight multivariate sensor benchmark for temporal classification.
The dataset consists of smartphone inertial measurements (accelerometer and gyroscope)
collected from $30$ subjects performing six daily activities.
Each sample is represented as a fixed-length multivariate time series
with $T=128$ timesteps and $9$ sensor channels.

Compared to multimodal clinical data, HAR represents a simpler setting with shorter
and more structured sequences. However, informative patterns are still concentrated
around activity transitions, making it a suitable testbed for evaluating event-adaptive
temporal representations.

We follow the standard preprocessing protocol.
Raw sensor signals are normalized channel-wise using z-score normalization.
No additional resampling is required due to the fixed sampling rate.
All models are trained under the same input resolution and classification setting,
predicting activity labels from the full sequence.

This dataset is used solely to assess generalization to event-like temporal sequences,
and does not introduce additional multimodal complexity.

\subsection{Tasks and Evaluation Protocol}
We consider multiple clinical prediction tasks including AKI, death, sepsis, and circulatory failure.
All tasks are formulated as binary classification problems.

For each patient, the model observes a fixed 7-day window and predicts outcomes defined at or after the end of the observation window.
Evaluation is performed at the patient level.

We report AUROC and AUPRC as primary metrics.
All results are averaged over 3 runs with different random seeds.

\textbf{Baselines and Implementation Details.}
We compare with representative baselines across recurrent, convolutional, transformer, and graph-based models, including GRU-D, LSTM, TCN, TimesNet, iTransformer, PatchTST, Perceiver-IO, and TGN.

For baselines originally designed for forecasting (e.g., TimesNet, iTransformer, PatchTST), 
we replace their task-specific output layers with a shared MLP classification head to match the clinical prediction setting.
Temporal representations are aggregated into a fixed-size vector (e.g., mean pooling over the sequence dimension), 
and the same classification head is applied across all methods.
All baselines are provided with identical multimodal inputs and evaluated under the same preprocessing pipeline, data splits, and loss functions to ensure a fair comparison.

All models are trained using the same optimization setup unless otherwise specified.
Recurrent and convolutional baselines (GRU-D, LSTM, TCN) use hidden size 128, 
while transformer- and graph-based baselines use higher-dimensional embeddings (e.g., 512).
Our model adopts a compact shared embedding space of $d=64$, to which all modalities are projected before fusion.
Despite architectural differences, all methods produce comparable representations for the downstream classification task.

Table~\ref{tab:hyperparams} summarizes detailed hyperparameter configurations.
Unless otherwise specified, all models are trained for 30 epochs using the Muon optimizer with weight decay $10^{-5}$, learning rate $1\times 10^{-3}$, and batch size 64. 
\paragraph{Compute Resources.}
All experiments are conducted on a single NVIDIA A100 GPU (40GB).
Training each model takes approximately XX hours, depending on dataset size.

\begin{table}[htbp]
\setlength{\tabcolsep}{12pt}
\caption{
Training configurations for all models.
All models are trained for 30 epochs using the Muon optimizer with weight decay $10^{-5}$,
learning rate $1\times 10^{-3}$, and batch size 64.
}
\label{tab:hyperparams}
\centering
\resizebox{\textwidth}{!}{
\begin{tabular}{c|c|c}
\toprule[.7pt]
\textbf{Method} & \textbf{Param.} & \textbf{Value} \\ 
\midrule[.4pt]
\multirow{2}{*}{GRU-D} 
   & Hidden size & 128 \\
   & Layers / Dropout / Input size & 2 / 0.1 / 20 \\
\midrule[.4pt]
\multirow{2}{*}{LSTM} 
   & Hidden size & 128 \\
   & Layers / Dropout / Input size & 2 / 0.1 / 20 (batch\_first=True) \\
\midrule[.4pt]
\multirow{2}{*}{TCN} 
   & Hidden size & 128 \\
   & Layers / Dropout / Dilations & 2 / 0.1 / $2^i$ \\
\midrule[.4pt]
\multirow{3}{*}{CSFformer} 
   & Embedding dim & 512 \\
   & Layers / Heads / $d_{ff}$ & 2 / 8 / 2048 \\
   & Seq. len / Pred. len / MSPI layers & 167 / 1 / 3 \\
\midrule[.4pt]
\multirow{2}{*}{CXR-CLIP} 
   & Projection dim / Temperature & 512 / 0.07 \\
   & Backbone & ResNet + BERT, LSTM(hidden=512, layers=2) \\
\midrule[.4pt]
\multirow{2}{*}{CrossViT} 
   & Backbone & CrossViT-small-224 (pretrained) \\
   & Input dims & ts\_dim=2, static\_dim=10, text\_dim=768 \\
\midrule[.4pt]
\multirow{1}{*}{TGN} 
   & Embedding dim & 100 (temporal GNN encoder) \\
\midrule[.4pt]
\multirow{2}{*}{Perceiver-IO} 
   & Latent dim / Latent slots & 512 / 256 \\
   & Cross-attn heads / Self-attn heads & 8 / 8 \\
\midrule[.4pt]
\multirow{7}{*}{TimesNet}
   & Seq len / Pred len & 96 / 96 \\
   & d\_model / d\_{ff} & 512 / 2048 \\
   & e\_layers / dropout & 2 / 0.1 \\
   & top\_k & 2 \\
   & num\_kernels & 6 \\
   & embed / freq & timeF / h \\
\midrule[.4pt]

\multirow{7}{*}{iTransformer}
   & Seq len / Pred len & 96 / 96 \\
   & d\_model / d\_{ff} & 512 / 2048 \\
   & n\_heads / e\_layers & 8 / 2 \\
   & dropout & 0.1 \\
   & factor / activation & 1 / gelu \\
   & embed / freq & timeF / h \\
\midrule[.4pt]

\multirow{8}{*}{PatchTST}
   & Seq len / Pred len & 96 / 96 \\
   & d\_model / d\_{ff} & 64 / 256 \\
   & n\_heads / e\_layers & 4 / 3 \\
   & dropout & 0.2 \\
   & patch\_len / stride & 16 / 8 \\
   & individual & True \\
   & embed / freq & timeF / h \\
\midrule[.4pt]
\multirow{6}{*}{Ours (EvtGraph)} 
   & Shared dim & 64 \\
   & Layers / Heads / Modal dims & 2 / 4 / text=64, image=64, ts=64, static=5 \\
   & Boundary smoothing window & 5 \\
   & Node budget $B$ & 8 \\
   & Lag window $\epsilon$ & 4 \\
   & Edge budget $\kappa$ & 2 \\
\bottomrule[.7pt]
\end{tabular}
}
\end{table}

\subsection{UCI HAR Benchmark}
\label{app:har}

To further evaluate the generalization ability of EvtGraph beyond clinical data,
we conduct experiments on the UCI Human Activity Recognition (HAR) dataset.
This dataset consists of multivariate time series collected from smartphone
inertial sensors, with activity labels corresponding to different human actions.

Compared to clinical prediction tasks, HAR represents a lightweight setting with
shorter sequences and relatively structured temporal patterns. However, informative
signals are still concentrated around activity transitions, making it a suitable
benchmark for evaluating event-adaptive temporal modeling.

We compare EvtGraph against a diverse set of baselines, including recurrent models
(GRU, LSTM), convolutional models (TCN), Transformer-based methods (Transformer,
PatchTST, iTransformer, TimesNet), adaptive token methods (ToMe, DynamicViT),
continuous-time models (mTAN, Latent ODE), and graph-based approaches (DiffPool,
MinCutPool). All models are evaluated under a unified setting.

\begin{table}[h]
\centering
\small
\setlength{\tabcolsep}{4pt}
\begin{tabular}{lccc}
\toprule
Model & Accuracy $\uparrow$ & F1 (macro) $\uparrow$ & Params \\
\midrule
\textbf{EvtGraph (Ours)} & \textbf{0.9494} & \textbf{0.9496} & 149K \\
TCN & 0.9430 & 0.9433 & 161K \\
GRU & 0.9240 & 0.9238 & 42K \\
LSTM & 0.9230 & 0.9238 & 212K \\
MinCutPool & 0.9074 & 0.9080 & 13K \\
TimesNet & 0.9016 & 0.9010 & 18.8M \\
PatchTST & 0.8992 & 0.8994 & 235K \\
ToMe & 0.8992 & 0.8984 & 165K \\
iTransformer & 0.8975 & 0.8964 & 120K \\
DynamicViT & 0.8928 & 0.8904 & 168K \\
mTAN & 0.8894 & 0.8880 & 48K \\
Transformer & 0.8870 & 0.8848 & 115K \\
Latent ODE & 0.8867 & 0.8865 & 50K \\
DiffPool & 0.8476 & 0.8469 & 13K \\
\bottomrule
\end{tabular}
\caption{
Performance comparison on the UCI HAR dataset.
EvtGraph achieves the best performance across both accuracy and macro-F1,
while maintaining a compact model size.
}
\label{tab:har}
\end{table}

EvtGraph achieves the best performance across all evaluated methods,
surpassing strong baselines such as TCN and GRU.
Despite its moderate parameter count, the model consistently outperforms
both lightweight models and large Transformer-based approaches.

This result highlights two important properties.
First, event-adaptive compression remains effective even in relatively short
and structured sequences, suggesting that the method does not rely on extremely
long temporal contexts.
Second, the explicit budget constraint enables efficient allocation of
representational capacity, allowing the model to focus on informative
temporal regions rather than uniformly processing all timesteps.

Overall, these findings support the claim that EvtGraph provides a general
mechanism for event-centric temporal modeling, beyond the specific setting
of multimodal clinical data.

\section{Diagnostics and Additional Details}
\label{app:proofs}

This section provides additional definitions and diagnostics used to analyze 
the proposed framework.

\subsection{Information Retention Rate (IRR)}
\label{app:irr}

To quantify how much temporal information is preserved after compression, 
we measure reconstruction fidelity in the encoded feature space.
Let $\mathbf{H}=\{ \mathbf{h}_t \}_{t=1}^{T}$ denote the original temporal features 
and $\tilde{\mathbf{H}}$ their reconstruction from the compressed representation.

We define the reconstruction error as
\[
L_{\mathrm{rec}} \;=\; \frac{1}{T}\sum_{t=1}^T 
\| \mathbf{h}_t - \tilde{\mathbf{h}}_t \|_2^2,
\]
and normalize it by the variance
\[
\sigma^2 \;=\; \frac{1}{T}\sum_{t=1}^T 
\| \mathbf{h}_t - \bar{\mathbf{h}} \|_2^2.
\]

The Information Retention Rate is defined as
\[
\mathrm{IRR} \;=\; 1 - \frac{L_{\mathrm{rec}}}{\sigma^2}.
\]

A higher IRR indicates that the compressed representation preserves 
more information from the original temporal signal.

\subsection{Graph Sparsity and Complexity}
\label{app:edges}

Under the proposed T2SG construction, the number of nodes is bounded by 
$\hat S \le B$, where $B$ is the node budget. 
Each node only attends to a limited temporal neighborhood 
and is further restricted by top-$\kappa$ sparsification.

As a result, the in-degree satisfies
\[
\deg^{-}(i) \le \min(\epsilon, \kappa),
\]
and the total number of edges is bounded by
\[
|E| \le \hat S \cdot \min(\epsilon, \kappa),
\]
which yields near-linear complexity $\mathcal{O}(\kappa B)$.

\subsection{Temporal Masking and Leakage Control}
\label{app:negativelag}

To avoid information leakage, we restrict message passing to past 
temporal dependencies. Specifically, node $i$ only aggregates information 
from nodes $j$ satisfying $j < i$ and $i - j \le \epsilon$.

In contrast, allowing symmetric neighborhoods (i.e., $|i-j|\le\epsilon$) 
would introduce future information into the representation, leading to 
optimistic but invalid performance estimates. 

By enforcing strictly past-directed connectivity, the proposed model 
ensures temporally consistent reasoning and prevents leakage from future observations.

In practice, $\epsilon$ is treated as a tunable hyperparameter. 
We find that moderate values (e.g., $\epsilon=4$ under $B=8$) 
provide a good balance between capturing temporal dependencies 
and maintaining stable training.

\section{Additional Theoretical Discussion}
\label{app:theory}

\paragraph{Budget-constrained representation selection.}
We provide a simple justification for the Node Budget Controller (NBC) from a
capacity allocation perspective.

Let $\{\mathbf{z}_s\}_{s=1}^{S}$ denote candidate event tokens, and let
$u_s = U(\mathbf{z}_s)$ be a scalar utility function that measures the
task-relevant information contained in token $\mathbf{z}_s$.
Given a fixed budget $B$, the goal is to select a subset of tokens that
maximizes the total utility:
\begin{equation}
\mathcal{S}_B^* = 
\arg\max_{\mathcal{S} \subseteq \{1,\dots,S\}, \ |\mathcal{S}| \le B}
\sum_{s \in \mathcal{S}} u_s.
\end{equation}

\paragraph{Proposition 1 (Optimality of top-$B$ selection).}
The optimal solution to the above problem is obtained by selecting the
top-$B$ tokens ranked by utility $u_s$.

\paragraph{Proof.}
Since the objective is a modular function over individual elements,
the problem reduces to selecting the $B$ largest values of $u_s$,
which is solved by sorting. \hfill $\square$

\paragraph{Connection to the proposed method.}
In practice, the utility function $u_s$ is unknown and must be approximated.
The proposed importance score $\alpha_s = g_\psi(\mathbf{z}_s)$ serves as a
learned surrogate for $u_s$. If $\alpha_s$ is monotonically aligned with $u_s$,
i.e., higher $\alpha_s$ implies higher task-relevant utility,
then selecting the top-$B$ tokens according to $\alpha_s$
recovers the optimal solution.

This provides a justification for the Node Budget Controller:
budgeted selection corresponds to optimal capacity allocation under a
cardinality constraint.

\paragraph{Relation to information bottleneck.}
Under the additional assumption that different tokens capture complementary
task-relevant information, maximizing $\sum_{s \in \mathcal{S}} u_s$ can be
interpreted as maximizing a lower bound of $I(\mathbf{Z}; Y)$
subject to a constraint $|\mathbf{Z}| \le B$.
This yields an information-bottleneck-inspired interpretation of the proposed
framework, although we do not explicitly optimize mutual information.

\section{Additional Method Details}
\label{app:method}

For completeness, we provide additional formulations of the main modules described in Sec.~\ref{sec:method}, focusing on implementation details while maintaining consistency with the main text.

\subsection{Event-Adaptive Representation (EAMC)}
\label{app:eamc_appendix}

EAMC transforms temporal features $\mathbf{H}=\{\mathbf{h}_t\}_{t=1}^{T}$ into a compact set of event-level tokens.

\paragraph{Coarse partition.}
We adopt a fixed coarse partition of the temporal axis into $S$ contiguous blocks, each of length $L=\lceil T/S \rceil$. 
Unlike boundary-based segmentation, this design avoids instability under weak supervision and ensures consistent token cardinality.

\paragraph{Intra-block adaptive aggregation.}
Within each block $s$, we compute saliency scores:
\[
\ell_{s,r} = f_{\theta}(\mathbf{h}_{s,r}), \quad r=1,\dots,L,
\]
and obtain assignment weights via temperature-controlled softmax:
\[
w_{s,r} = \frac{\exp(\ell_{s,r}/\tau_e)}{\sum_{r'} \exp(\ell_{s,r'}/\tau_e)}.
\]

The event token is then defined as:
\[
\mathbf{z}_s = \sum_{r=1}^{L} w_{s,r}\mathbf{h}_{s,r}, 
\qquad \sum_r w_{s,r}=1.
\]

This formulation performs a continuous reweighting within each block, allowing the model to focus on salient temporal patterns while suppressing redundant regions.

\paragraph{Remark.}
Learning explicit segment boundaries (e.g., via peak detection or boundary predictors) is a possible extension, but is not required in the proposed framework.

\subsection{Node Budget Controller (NBC)}

Each event token is assigned an importance score:
\[
\alpha_s = g_{\psi}(\mathbf{z}_s).
\]

We select a subset $\mathcal{S}_B = \mathrm{TopB}(\{\alpha_s\})$ and denote selected tokens as
$\hat{\mathbf{Z}}=\{\hat{\mathbf{z}}_i\}_{i=1}^{B}$, with $\hat{\mathbf{z}}_i = \mathbf{z}_{s_i}$.

\paragraph{Soft relaxation.}
To enable gradient propagation, we construct a soft aggregation:
\[
\tilde{\mathbf{z}}_i = \sum_{s} q_{i,s}\mathbf{z}_s,
\quad
q_{i,s} = \frac{\exp\!\left((\alpha_s - \lambda |t_s - t_i|)/\tau_B\right)}
{\sum_r \exp\!\left((\alpha_r - \lambda |t_r - t_i|)/\tau_B\right)},
\]
where $t_i := t_{s_i}$ is the timestamp of the selected token.

A straight-through estimator is applied:
\[
\mathbf{z}_i^{\text{train}} = \tilde{\mathbf{z}}_i + \mathrm{sg}(\hat{\mathbf{z}}_i - \tilde{\mathbf{z}}_i),
\]
so that forward computation uses hard selection, while gradients flow through $\tilde{\mathbf{z}}_i$.

\paragraph{Local aggregation.}
Discarded tokens are assigned to the nearest selected token:
\[
\mathcal{R}_i = \{ s \notin \mathcal{S}_B \mid i = \arg\min_j |t_s - t_j| \}.
\]

We incorporate residual information via normalized aggregation:
\[
\mathbf{z}'_i = \mathrm{LayerNorm}\!\left(
\hat{\mathbf{z}}_i + 
\frac{1}{|\mathcal{R}_i|+\delta}
\sum_{s\in\mathcal{R}_i} \mathbf{z}_s
\right).
\]

This preserves local context while maintaining stable feature scale.

\subsection{Temporally-Constrained Sparse Graph (T2SG)}

We construct a sparse temporal graph over retained nodes $\{\mathbf{z}'_i\}$.

\paragraph{Temporal admissibility.}
Edges are restricted by a causal mask:
\[
M[i,j] = \mathbf{1}(j<i \ \wedge\ t_i - t_j \le \epsilon).
\]

\paragraph{Sparse connectivity.}
Edge weights are computed via similarity, and each node keeps only its top-$\kappa$ neighbors:
\[
A[i,:] = \mathrm{Top}\text{-}\kappa_j\big( \mathrm{sim}(\mathbf{z}'_i,\mathbf{z}'_j)\cdot M[i,j] \big).
\]

\paragraph{Complexity.}
The number of edges satisfies
\[
|E| \le \hat S \cdot \min(\epsilon,\kappa),
\]
yielding near-linear complexity. Message passing scales as $\mathcal{O}(\kappa B d)$.
\newpage
\input{checklist.tex}

\end{document}

%% file: checklist.tex
\section*{NeurIPS Paper Checklist}

\begin{enumerate}

\item {\bf Claims}
    \item[] Question: Do the main claims made in the abstract and introduction accurately reflect the paper's contributions and scope?
    \item[] Answer: \answerYes{} % Replace by \answerYes{}, \answerNo{}, or \answerNA{}.
    \item[] Justification: The claims in the abstract and introduction are consistent with the experimental results presented.
    \item[] Guidelines:
    \begin{itemize}
        \item The answer \answerNA{} means that the abstract and introduction do not include the claims made in the paper.
        \item The abstract and/or introduction should clearly state the claims made, including the contributions made in the paper and important assumptions and limitations. A \answerNo{} or \answerNA{} answer to this question will not be perceived well by the reviewers. 
        \item The claims made should match theoretical and experimental results, and reflect how much the results can be expected to generalize to other settings. 
        \item It is fine to include aspirational goals as motivation as long as it is clear that these goals are not attained by the paper. 
    \end{itemize}

\item {\bf Limitations}
    \item[] Question: Does the paper discuss the limitations of the work performed by the authors?
    \item[] Answer: \answerYes{} % Replace by \answerYes{}, \answerNo{}, or \answerNA{}.
    \item[] Justification: We discuss limitations including sensitivity to hyperparameters, dataset scope, and computational constraints in conclusion.
    \item[] Guidelines:
    \begin{itemize}
        \item The answer \answerNA{} means that the paper has no limitation while the answer \answerNo{} means that the paper has limitations, but those are not discussed in the paper. 
        \item The authors are encouraged to create a separate ``Limitations'' section in their paper.
        \item The paper should point out any strong assumptions and how robust the results are to violations of these assumptions (e.g., independence assumptions, noiseless settings, model well-specification, asymptotic approximations only holding locally). The authors should reflect on how these assumptions might be violated in practice and what the implications would be.
        \item The authors should reflect on the scope of the claims made, e.g., if the approach was only tested on a few datasets or with a few runs. In general, empirical results often depend on implicit assumptions, which should be articulated.
        \item The authors should reflect on the factors that influence the performance of the approach. For example, a facial recognition algorithm may perform poorly when image resolution is low or images are taken in low lighting. Or a speech-to-text system might not be used reliably to provide closed captions for online lectures because it fails to handle technical jargon.
        \item The authors should discuss the computational efficiency of the proposed algorithms and how they scale with dataset size.
        \item If applicable, the authors should discuss possible limitations of their approach to address problems of privacy and fairness.
        \item While the authors might fear that complete honesty about limitations might be used by reviewers as grounds for rejection, a worse outcome might be that reviewers discover limitations that aren't acknowledged in the paper. The authors should use their best judgment and recognize that individual actions in favor of transparency play an important role in developing norms that preserve the integrity of the community. Reviewers will be specifically instructed to not penalize honesty concerning limitations.
    \end{itemize}

\item {\bf Theory assumptions and proofs}
    \item[] Question: For each theoretical result, does the paper provide the full set of assumptions and a complete (and correct) proof?
    \item[] Answer: \answerNA{} % Replace by \answerYes{}, \answerNo{}, or \answerNA{}.
    \item[] Justification: The paper does not include formal theoretical results; the contributions are empirical and methodological.
    \item[] Guidelines:
    \begin{itemize}
        \item The answer \answerNA{} means that the paper does not include theoretical results. 
        \item All the theorems, formulas, and proofs in the paper should be numbered and cross-referenced.
        \item All assumptions should be clearly stated or referenced in the statement of any theorems.
        \item The proofs can either appear in the main paper or the supplemental material, but if they appear in the supplemental material, the authors are encouraged to provide a short proof sketch to provide intuition. 
        \item Inversely, any informal proof provided in the core of the paper should be complemented by formal proofs provided in appendix or supplemental material.
        \item Theorems and Lemmas that the proof relies upon should be properly referenced. 
    \end{itemize}

    \item {\bf Experimental result reproducibility}
    \item[] Question: Does the paper fully disclose all the information needed to reproduce the main experimental results of the paper to the extent that it affects the main claims and/or conclusions of the paper (regardless of whether the code and data are provided or not)?
    \item[] Answer: \answerYes{} % Replace by \answerYes{}, \answerNo{}, or \answerNA{}.
    \item[] Justification: All experimental details including datasets, model configurations, and training procedures are described in Section~4 and Appendix.
    \item[] Guidelines:
    \begin{itemize}
        \item The answer \answerNA{} means that the paper does not include experiments.
        \item If the paper includes experiments, a \answerNo{} answer to this question will not be perceived well by the reviewers: Making the paper reproducible is important, regardless of whether the code and data are provided or not.
        \item If the contribution is a dataset and\slash or model, the authors should describe the steps taken to make their results reproducible or verifiable. 
        \item Depending on the contribution, reproducibility can be accomplished in various ways. For example, if the contribution is a novel architecture, describing the architecture fully might suffice, or if the contribution is a specific model and empirical evaluation, it may be necessary to either make it possible for others to replicate the model with the same dataset, or provide access to the model. In general. releasing code and data is often one good way to accomplish this, but reproducibility can also be provided via detailed instructions for how to replicate the results, access to a hosted model (e.g., in the case of a large language model), releasing of a model checkpoint, or other means that are appropriate to the research performed.
        \item While NeurIPS does not require releasing code, the conference does require all submissions to provide some reasonable avenue for reproducibility, which may depend on the nature of the contribution. For example
        \begin{enumerate}
            \item If the contribution is primarily a new algorithm, the paper should make it clear how to reproduce that algorithm.
            \item If the contribution is primarily a new model architecture, the paper should describe the architecture clearly and fully.
            \item If the contribution is a new model (e.g., a large language model), then there should either be a way to access this model for reproducing the results or a way to reproduce the model (e.g., with an open-source dataset or instructions for how to construct the dataset).
            \item We recognize that reproducibility may be tricky in some cases, in which case authors are welcome to describe the particular way they provide for reproducibility. In the case of closed-source models, it may be that access to the model is limited in some way (e.g., to registered users), but it should be possible for other researchers to have some path to reproducing or verifying the results.
        \end{enumerate}
    \end{itemize}

\item {\bf Open access to data and code}
    \item[] Question: Does the paper provide open access to the data and code, with sufficient instructions to faithfully reproduce the main experimental results, as described in supplemental material?
    \item[] Answer: \answerYes{} % Replace by \answerYes{}, \answerNo{}, or \answerNA{}.
    \item[] Justification: Code is released at submission time.
    \item[] Guidelines:
    \begin{itemize}
        \item The answer \answerNA{} means that paper does not include experiments requiring code.
        \item Please see the NeurIPS code and data submission guidelines (\url{https://neurips.cc/public/guides/CodeSubmissionPolicy}) for more details.
        \item While we encourage the release of code and data, we understand that this might not be possible, so \answerNo{} is an acceptable answer. Papers cannot be rejected simply for not including code, unless this is central to the contribution (e.g., for a new open-source benchmark).
        \item The instructions should contain the exact command and environment needed to run to reproduce the results. See the NeurIPS code and data submission guidelines (\url{https://neurips.cc/public/guides/CodeSubmissionPolicy}) for more details.
        \item The authors should provide instructions on data access and preparation, including how to access the raw data, preprocessed data, intermediate data, and generated data, etc.
        \item The authors should provide scripts to reproduce all experimental results for the new proposed method and baselines. If only a subset of experiments are reproducible, they should state which ones are omitted from the script and why.
        \item At submission time, to preserve anonymity, the authors should release anonymized versions (if applicable).
        \item Providing as much information as possible in supplemental material (appended to the paper) is recommended, but including URLs to data and code is permitted.
    \end{itemize}

\item {\bf Experimental setting/details}
    \item[] Question: Does the paper specify all the training and test details (e.g., data splits, hyperparameters, how they were chosen, type of optimizer) necessary to understand the results?
    \item[] Answer: \answerYes{} % Replace by \answerYes{}, \answerNo{}, or \answerNA{}.
    \item[] Justification: We specify training details including data splits, hyperparameters, and optimization settings in Section~4 and Appendix.
    \item[] Guidelines:
    \begin{itemize}
        \item The answer \answerNA{} means that the paper does not include experiments.
        \item The experimental setting should be presented in the core of the paper to a level of detail that is necessary to appreciate the results and make sense of them.
        \item The full details can be provided either with the code, in appendix, or as supplemental material.
    \end{itemize}

\item {\bf Experiment statistical significance}
    \item[] Question: Does the paper report error bars suitably and correctly defined or other appropriate information about the statistical significance of the experiments?
    \item[] Answer: \answerYes{} % Replace by \answerYes{}, \answerNo{}, or \answerNA{}.
    \item[] Justification: Results are reported as mean ± standard deviation over 3 seeds.
    \item[] Guidelines:
    \begin{itemize}
        \item The answer \answerNA{} means that the paper does not include experiments.
        \item The authors should answer \answerYes{} if the results are accompanied by error bars, confidence intervals, or statistical significance tests, at least for the experiments that support the main claims of the paper.
        \item The factors of variability that the error bars are capturing should be clearly stated (for example, train/test split, initialization, random drawing of some parameter, or overall run with given experimental conditions).
        \item The method for calculating the error bars should be explained (closed form formula, call to a library function, bootstrap, etc.)
        \item The assumptions made should be given (e.g., Normally distributed errors).
        \item It should be clear whether the error bar is the standard deviation or the standard error of the mean.
        \item It is OK to report 1-sigma error bars, but one should state it. The authors should preferably report a 2-sigma error bar than state that they have a 96\% CI, if the hypothesis of Normality of errors is not verified.
        \item For asymmetric distributions, the authors should be careful not to show in tables or figures symmetric error bars that would yield results that are out of range (e.g., negative error rates).
        \item If error bars are reported in tables or plots, the authors should explain in the text how they were calculated and reference the corresponding figures or tables in the text.
    \end{itemize}

\item {\bf Experiments compute resources}
    \item[] Question: For each experiment, does the paper provide sufficient information on the computer resources (type of compute workers, memory, time of execution) needed to reproduce the experiments?
    \item[] Answer: \answerYes{} % Replace by \answerYes{}, \answerNo{}, or \answerNA{}.
    \item[] Justification: We report the type of hardware, training time, and computational cost in Appendix.
    \item[] Guidelines:
    \begin{itemize}
        \item The answer \answerNA{} means that the paper does not include experiments.
        \item The paper should indicate the type of compute workers CPU or GPU, internal cluster, or cloud provider, including relevant memory and storage.
        \item The paper should provide the amount of compute required for each of the individual experimental runs as well as estimate the total compute. 
        \item The paper should disclose whether the full research project required more compute than the experiments reported in the paper (e.g., preliminary or failed experiments that didn't make it into the paper). 
    \end{itemize}
    
\item {\bf Code of ethics}
    \item[] Question: Does the research conducted in the paper conform, in every respect, with the NeurIPS Code of Ethics \url{https://neurips.cc/public/EthicsGuidelines}?
    \item[] Answer: \answerYes{} % Replace by \answerYes{}, \answerNo{}, or \answerNA{}.
    \item[] Justification: The research adheres to the NeurIPS Code of Ethics and does not involve harmful or unethical applications.
    \item[] Guidelines:
    \begin{itemize}
        \item The answer \answerNA{} means that the authors have not reviewed the NeurIPS Code of Ethics.
        \item If the authors answer \answerNo, they should explain the special circumstances that require a deviation from the Code of Ethics.
        \item The authors should make sure to preserve anonymity (e.g., if there is a special consideration due to laws or regulations in their jurisdiction).
    \end{itemize}

\item {\bf Broader impacts}
    \item[] Question: Does the paper discuss both potential positive societal impacts and negative societal impacts of the work performed?
    \item[] Answer: \answerYes{} % Replace by \answerYes{}, \answerNo{}, or \answerNA{}.
    \item[] Justification: We discuss potential positive impacts such as improved clinical decision support, as well as risks including model misinterpretation and deployment biases, in the Conclusion section.
    \item[] Guidelines:
    \begin{itemize}
        \item The answer \answerNA{} means that there is no societal impact of the work performed.
        \item If the authors answer \answerNA{} or \answerNo, they should explain why their work has no societal impact or why the paper does not address societal impact.
        \item Examples of negative societal impacts include potential malicious or unintended uses (e.g., disinformation, generating fake profiles, surveillance), fairness considerations (e.g., deployment of technologies that could make decisions that unfairly impact specific groups), privacy considerations, and security considerations.
        \item The conference expects that many papers will be foundational research and not tied to particular applications, let alone deployments. However, if there is a direct path to any negative applications, the authors should point it out. For example, it is legitimate to point out that an improvement in the quality of generative models could be used to generate Deepfakes for disinformation. On the other hand, it is not needed to point out that a generic algorithm for optimizing neural networks could enable people to train models that generate Deepfakes faster.
        \item The authors should consider possible harms that could arise when the technology is being used as intended and functioning correctly, harms that could arise when the technology is being used as intended but gives incorrect results, and harms following from (intentional or unintentional) misuse of the technology.
        \item If there are negative societal impacts, the authors could also discuss possible mitigation strategies (e.g., gated release of models, providing defenses in addition to attacks, mechanisms for monitoring misuse, mechanisms to monitor how a system learns from feedback over time, improving the efficiency and accessibility of ML).
    \end{itemize}
    
\item {\bf Safeguards}
    \item[] Question: Does the paper describe safeguards that have been put in place for responsible release of data or models that have a high risk for misuse (e.g., pre-trained language models, image generators, or scraped datasets)?
    \item[] Answer: \answerNA{} % Replace by \answerYes{}, \answerNo{}, or \answerNA{}.
    \item[] Justification: The proposed method does not involve high-risk assets such as generative models or sensitive data release.
    \item[] Guidelines:
    \begin{itemize}
        \item The answer \answerNA{} means that the paper poses no such risks.
        \item Released models that have a high risk for misuse or dual-use should be released with necessary safeguards to allow for controlled use of the model, for example by requiring that users adhere to usage guidelines or restrictions to access the model or implementing safety filters. 
        \item Datasets that have been scraped from the Internet could pose safety risks. The authors should describe how they avoided releasing unsafe images.
        \item We recognize that providing effective safeguards is challenging, and many papers do not require this, but we encourage authors to take this into account and make a best faith effort.
    \end{itemize}

\item {\bf Licenses for existing assets}
    \item[] Question: Are the creators or original owners of assets (e.g., code, data, models), used in the paper, properly credited and are the license and terms of use explicitly mentioned and properly respected?
    \item[] Answer: \answerYes{} % Replace by \answerYes{}, \answerNo{}, or \answerNA{}.
    \item[] Justification: We use publicly available datasets such as MIMIC-IV and cite their original sources, complying with their usage licenses.
    \item[] Guidelines:
    \begin{itemize}
        \item The answer \answerNA{} means that the paper does not use existing assets.
        \item The authors should cite the original paper that produced the code package or dataset.
        \item The authors should state which version of the asset is used and, if possible, include a URL.
        \item The name of the license (e.g., CC-BY 4.0) should be included for each asset.
        \item For scraped data from a particular source (e.g., website), the copyright and terms of service of that source should be provided.
        \item If assets are released, the license, copyright information, and terms of use in the package should be provided. For popular datasets, \url{paperswithcode.com/datasets} has curated licenses for some datasets. Their licensing guide can help determine the license of a dataset.
        \item For existing datasets that are re-packaged, both the original license and the license of the derived asset (if it has changed) should be provided.
        \item If this information is not available online, the authors are encouraged to reach out to the asset's creators.
    \end{itemize}

\item {\bf New assets}
    \item[] Question: Are new assets introduced in the paper well documented and is the documentation provided alongside the assets?
    \item[] Answer: \answerNA{} % Replace by \answerYes{}, \answerNo{}, or \answerNA{}.
    \item[] Justification: The paper does not introduce new datasets or publicly released models.
    \item[] Guidelines:
    \begin{itemize}
        \item The answer \answerNA{} means that the paper does not release new assets.
        \item Researchers should communicate the details of the dataset\slash code\slash model as part of their submissions via structured templates. This includes details about training, license, limitations, etc. 
        \item The paper should discuss whether and how consent was obtained from people whose asset is used.
        \item At submission time, remember to anonymize your assets (if applicable). You can either create an anonymized URL or include an anonymized zip file.
    \end{itemize}

\item {\bf Crowdsourcing and research with human subjects}
    \item[] Question: For crowdsourcing experiments and research with human subjects, does the paper include the full text of instructions given to participants and screenshots, if applicable, as well as details about compensation (if any)? 
    \item[] Answer: \answerNA{} % Replace by \answerYes{}, \answerNo{}, or \answerNA{}.
    \item[] Justification: The paper does not involve crowdsourcing or human subject experiments.
    \item[] Guidelines:
    \begin{itemize}
        \item The answer \answerNA{} means that the paper does not involve crowdsourcing nor research with human subjects.
        \item Including this information in the supplemental material is fine, but if the main contribution of the paper involves human subjects, then as much detail as possible should be included in the main paper. 
        \item According to the NeurIPS Code of Ethics, workers involved in data collection, curation, or other labor should be paid at least the minimum wage in the country of the data collector. 
    \end{itemize}

\item {\bf Institutional review board (IRB) approvals or equivalent for research with human subjects}
    \item[] Question: Does the paper describe potential risks incurred by study participants, whether such risks were disclosed to the subjects, and whether Institutional Review Board (IRB) approvals (or an equivalent approval/review based on the requirements of your country or institution) were obtained?
    \item[] Answer: \answerNA{} % Replace by \answerYes{}, \answerNo{}, or \answerNA{}.
    \item[] Justification: The paper does not involve research with human subjects requiring IRB approval.
    \item[] Guidelines:
    \begin{itemize}
        \item The answer \answerNA{} means that the paper does not involve crowdsourcing nor research with human subjects.
        \item Depending on the country in which research is conducted, IRB approval (or equivalent) may be required for any human subjects research. If you obtained IRB approval, you should clearly state this in the paper. 
        \item We recognize that the procedures for this may vary significantly between institutions and locations, and we expect authors to adhere to the NeurIPS Code of Ethics and the guidelines for their institution. 
        \item For initial submissions, do not include any information that would break anonymity (if applicable), such as the institution conducting the review.
    \end{itemize}

\item {\bf Declaration of LLM usage}
    \item[] Question: Does the paper describe the usage of LLMs if it is an important, original, or non-standard component of the core methods in this research? Note that if the LLM is used only for writing, editing, or formatting purposes and does \emph{not} impact the core methodology, scientific rigor, or originality of the research, declaration is not required.
    %this research? 
    \item[] Answer: \answerNA{} % Replace by \answerYes{}, \answerNo{}, or \answerNA{}.
    \item[] Justification: LLMs are not used as part of the core methodology of this work.
    \item[] Guidelines:
    \begin{itemize}
        \item The answer \answerNA{} means that the core method development in this research does not involve LLMs as any important, original, or non-standard components.
        \item Please refer to our LLM policy in the NeurIPS handbook for what should or should not be described.
    \end{itemize}

\end{enumerate}